\documentclass[11pt,twopage,runningheads]{llncs}

\usepackage{subfigure}
\usepackage{graphicx}
\usepackage{times}
\usepackage{tikz}

\begin{document}

\newcommand{\todo}{\fbox{ To Do }}
\newcommand{\todox}[1]{\fbox{ #1 }}
\newcommand*\circled[1]{\tikz[baseline=(char.base)]{
            \node[shape=circle,draw,inner sep=2pt] (char) {#1};}}
\newcommand{\procphase}[2]{\emph{#1}~\circled{#2}}

\newcommand{\prjplan}{\procphase{Project Planning and Initiation}{1}}
\newcommand{\ethicsreviewi}{\procphase{Ethics Review I}{2}}
\newcommand{\requirementselicitation}{\procphase{Requirements Elicitation}{3}}
\newcommand{\dataacquisition}{\procphase{Data Acquisition}{4}}
\newcommand{\feasibilitystudy}{\procphase{Feasibility Study}{5}}
\newcommand{\evaldesign}{\procphase{Evaluation Design}{6}}
\newcommand{\preproc}{\procphase{Data Pre-Processing and Cleansing}{7}}
\newcommand{\golddataannotation}{\procphase{Gold Data Annotation}{8}}
\newcommand{\sysdesign}{\procphase{System Architecture Design}{9}}
\newcommand{\framing}{\procphase{Choice of Machine Learning Classifier/Rule Formalism}{10}}
\newcommand{\sysimplementation}{\procphase{System Implementation}{11}}
\newcommand{\systesting}{\procphase{System Testing}{12}}

\newcommand{\features}{\procphase{Feature Design and Implementation}{13}}
\newcommand{\quantevali}{\procphase{Quantitative Evaluation I}{14}}
\newcommand{\patentingpub}{\procphase{Patenting and Publication}{15}}
\newcommand{\finalreport}{\procphase{Final Report Authoring}{16}}
\newcommand{\knowledgetransfer}{\procphase{Knowledge Transfer}{17}}
\newcommand{\acceptclosure}{\procphase{Acceptance and Closure}{18}}

\newcommand{\preannotate}{\procphase{Experimental Annotation of a Small Data Sample}{8.1}}
\newcommand{\authorguidlines}{\procphase{Authoring of Annotation Guidelines}{8.2}}
\newcommand{\annotate}{\procphase{Annotation of a Data Sample by Multiple Annotators}{8.3}}
\newcommand{\agreement}{\procphase{Computation of Inter-Annotator Agreement}{8.4}}
\newcommand{\adjudicate}{\procphase{Adjudication of Discrepancies}{8.5}}
\newcommand{\reviseguidelines}{\procphase{Revision of Annotation Guidelines}{8.6}}
\newcommand{\splitcorpus}{\procphase{Split Gold Data into Training Set, Dev-Test Set and Test Set}{8.7}}

\newcommand{\research}{\procphase{Research Project}{1-18}}
\newcommand{\ethicsreviewii}{\procphase{Ethics Review II}{19$\ast$}}
\newcommand{\deployment}{\procphase{Deployment}{20$\ast$}}
\newcommand{\monitoring}{\procphase{Monitoring}{21$\ast$}}
\newcommand{\quantevalii}{\procphase{Quantitative Evaluation II}{22$\ast$}}
\newcommand{\retraining}{\procphase{Model Re-Training}{23$\ast$}}

\title{Data to Value: An `Evaluation-First' Methodology for Natural Language Projects\thanks{This paper was written while the author spent time as a guest lecturer at the University of Zurich, Institute for Computational Linguistics. The author gratefully acknowledges the support of Martin Volk for extending the invitation to teach and for his hospitality, to Khalid Al-Kofahi (Thomson Reuters) for supporting this visit, and to Maria Fassli for the invitation to present a preliminary version at the Essex University Big Data Summer School. Thanks to Beatriz De La Iglesia for useful discussions.}}
\toctitle{D2V: A Methodology for NL Projects}

\author{Jochen L.~Leidner\inst{1,2}\\\email{leidner@acm.org}}
%
\authorrunning{Leidner}
\titlerunning{Data to Value (D2V)}
\institute{${}^1$ Coburg University of Applied Sciences, Friedrich-Streib-Str.~2, 96450 Coburg, Germany.\\${}^2$ University of Sheffield, Regents Court, 211 Portobelloe, Sheffield S1 5DP, England, United Kindgom.}

\date{}

\maketitle

\begin{abstract}
  \sloppypar{}
  Big data, i.e.~collecting, storing and processing of data at scale, has
  recently been possible due to the arrival of clusters of commodity computers
  powered by application-level distributed parallel operating systems like
  HDFS/Hadoop/Spark, and such infrastructures have revolutionized data mining
  at scale. For data mining project to succeed more consistently, some
  methodologies were developed (e.g.~CRISP-DM, SEMMA, KDD), but these do not
  account for (1) very large scales of processing, (2) dealing with 
  unstructured (textual) data (i.e., natural language processing also known as
  text analytics), and (3) non-technical considerations (e.g.~legal, ethical,
  project managerial aspects).\\
  To address these shortcomings, a new methodology, called \emph{``Data to
  Value'' (D2V)}, is introduced, which is guided by a detailed catalog of
  questions in order to avoid a disconnect of big data text analytics project
  team with the topic when facing rather abstract box-and-arrow diagrams
  commonly associated with methodologies.\\

  \textbf{Keywords.} Methodology; process model; supervised learning;
  natural language processing; data science; big data; unstructured
  data.
\end{abstract}

\section{Introduction} \label{sec:intro}

Engineering has been defined as ``the application of scientific,
economic, social, and practical knowledge in order to design, build,
and maintain structures, machines, devices, systems, materials and
processes.'' (Wikipedia) or the ``[t]he creative application of
scientific principles to design or develop structures, machines,
apparatus, or manufacturing processes, or works utilizing them singly
or in combination; or to construct or operate the same with full
cognizance of their design; or to forecast their behavior under
specific operating conditions; all as respects an intended function,
economics of operation or safety to life and property'' (American
Engineers' Council for Professional Development).

In language engineering, like in all engineering, we should adhere to
a \emph{principled approach} following \emph{best practices}
(methodology) and generate a \emph{predictable outcome}, namely: (1)
forecasting observed runtime, (2) forecasting memory requirements, (3)
forecasting delivery time, i.e.~a time at which a successful project
can be concluded; and (4) forecasting output quality (e.g.~F1-score).
At the time of writing, there is no theory that permits us to forecast
even one of these factors; in this paper, as a first step towards this
end, we present a methodology that addresses the point of delivery as
`predictable outcome'': our methodology aims to deliver projects more
consistently, and with less time wasted. This is achieved by a
finer-grained sequence of phases, and by additional guidance in the
form of guiding questions.
The unique characteristics of our methodology, named D2V (for ``Data
to Value'' process) are: it is \emph{evaluation-first}, which means that
the construction of a system and proceeding between the various phases
itself is driven by quantitative metrics; it is intended for projects
involving \emph{unstructured} data (text content), as it fills a gap
in the space of existing methodologies; it is \emph{question-informed},
as it benefits from a catalog of questions associated with particular
process phases.

The remainder of this paper is divided as follows.
Section~\ref{sec:related} briefly reviews the related work.
Section~\ref{sec:d2v} introduces the D2V methodology.
Section~\ref{sec:discussion} provides a discussion of its advantages
and limitations. Section~\ref{sec:conclusion} concludes with pointers
for future work.\\

\section{Related Work} \label{sec:related}

According to the PMI \cite{PMI:2013}, a project is completed successfully if
it is completed (1) on \emph{time} (2) on \emph{budget} (3) \emph{to
specification} and (4) with \emph{customer acceptance}. In the area of
software engineering, the waterfall model
and the agile model
have been the most popular
process models for constructing general software systems
\cite{Sommerville:2010}.

%

Working with data has a slightly different focus compared to traditional
software development: systems for mining rules, classifying documents, tagging
texts and extracting information are also software artifacts, but the software
co-evolves with various data sets and linguistic resources that are used by
it. While we are not aware of prior work on any methodologies specifically for
processing large quantities of text, in the context of developing data mining
projects, three popular methodologies have been developed, which shall be
reviewed here.

\sloppypar{}
The \textbf{KDD Process}~\cite{Fayyad-PiatetskyShapiro-Smyth:1996,Fayyad-PiatetskyShapiro-Smyth:1996:CACM,Debuse-etal:2001}
grew out of the Knowledge Discovery in Databases research community,
which in 1995 had its first of a series of workshops. It proposes a
sequence of five
\cite[p.~30-31]{Fayyad-PiatetskyShapiro-Smyth:1996:CACM} to nine
\cite[p.~29, Figure~1]{Fayyad-PiatetskyShapiro-Smyth:1996:CACM} steps
to get from raw data to knowledge: Selection, Pre-Processing,
Transformation, Data Mining and Interpretation/Evaluation.  Each of
these steps is seen as depending on the previous steps, yet its
proponents suggest a certain flexibility in applying the steps was
vital, and left to the discretion of the experienced researcher. For
example, at any stage could one consider going back to a any previous
stage and repeat the steps from there, and multiple iterations were
considered likely.  The initial step is to learn the application
domain. Then a target data set can be created by selecting it from all
available data. Raw data then gets cleaned up and pre-processed
(outlier/noise elimination), after which in a data reduction and
project phase, features helpful to the end goal are computed. A
function is then chosen (describing the purpose and identifying
whenther e.g.~classifaction or clustering can achieves it), and a
concrete method/algorithm is selected that implements the function.
Then the actual data mining (implementation and execution) step
happens, before the data is finally evaluated, interpreted and put to
use.

\sloppypar{}
The \textbf{SEMMA} methodology \cite{SAS:2013} was originally developed by the
private company SAS Institute, Inc. The name is an acronym, which stands ``for
Sample, Explore, Modify, Model, and Assess. SAS Enterprise Miner nodes ``are
arranged on tabs with the same names."~\cite[p.~5]{SAS:2013} In SEMMA, a cycle
with five process stages is applied: ``Sample -- This stage consists on
sampling the data by extracting a portion of a large data set big enough to
contain the significant information, yet small enough to manipulate quickly.
This stage is pointed out as being optional. Explore -- This stage consists on
the exploration of the data by searching for unanticipated trends and
anomalies in order to gain understanding and ideas. Modify --This stage
consists on the modification of the data by creating, selecting, and
transforming the variables to focus the model selection process.  Model --
This stage consists on modeling the data by allowing the software to search
automatically for a combination of data that reliably predicts a desired
outcome. Assess -- This stage consists on assessing the data by evaluating
the usefulness and reliability of the findings from the data mining process
and estimate how well it performs.'' \cite{Azevedo-Santos:2008:IADIS}.  

\sloppypar{}
The \textbf{CRISP-DM} methodology \cite{Chapman-etal:2000:TR,Shearer:2000:JDatWareh,Anand-etal:2007:ArtifIntellRev},
short for ``CRoss Industry Standard Process for Data Mining'', which
was developed by a consortium comprising DamilerChrysler, SPSS, NCR
and OHRA, can be looked at from four different levels of abstraction:
phases, generic tasks, specialized task and process instances. The
reference model defines the transitions between phases,
and a user guide
describes how-to information in more detail. It comprises six loose
phases: Business Understanding, Data Understanding, Data Preparation,
Modeling, Evaluation and Deployment.
See \cite{Kurgan-Musilek:2006:KnowEngRev,Anand-etal:2007:ArtifIntellRev,Azevedo-Santos:2008:IADIS} 
for more detail on the similarities and differences between KDD, SEMMA and
CRISP-DM.

%

\sloppypar{}
Marr's \textbf{Strategy Board}. \cite{Marr:2015} introduces a canvas-based
approach to guide big data projects, which focuses on the business aspects;
no technical guidance is provided by his model, but he does use a small number
of (6) guidance questions.

Here we would like to stress the shortcomings with regards to detail about a
number of areas of past work: First, working with unstructured data, textual
data in particular, is not specifically accommodated by either approach, despite
the fact that most non- transactional data in corporations comes in the form of
reports, white papers, slide decks, emails and other text-heavy formats.
Secondly, supervised learning, which is the approach of choice in scenario
where quality matters, is not specifically catered for by prior work. Third,
ethical questions, which are increasingly becoming more important in big data
work, are not part of  previously proposed processes. Fourth, the scale
aspects of big data influences the process and day-to-day work. Fifth and most
importantly, past work puts the evaluation phrase rather too late in the
overall process: for instance, in SEMMA the Assessment phase is the fourth of
five phases overall. In the author's view, this is the single most valuable
contribution of this paper, namely to strongly advocate an ``evaluation
first'' approach.

In contrast to past work, we present a process that is sensitive to ethical
concerns, accommodates big data projects using unstructured (textual) data
and supervised machine learning that typically goes with it, and which
requires textual annotation sub-processes.

\textbf{Evaluation design}: one of the earliest steps in a project following
our methodology is to
determine how the output of the system is to be evaluated.   This step is a
pecularity of the D2V process, and makes it   ``evaluation-first''. An
implenmentation of said evaluation's test   harness typically goes hand in
hand with the design.

%

\section{Data to Value (D2V)} \label{sec:d2v}

\begin{figure}
  \begin{center}
    {\includegraphics[width=.9\textwidth]{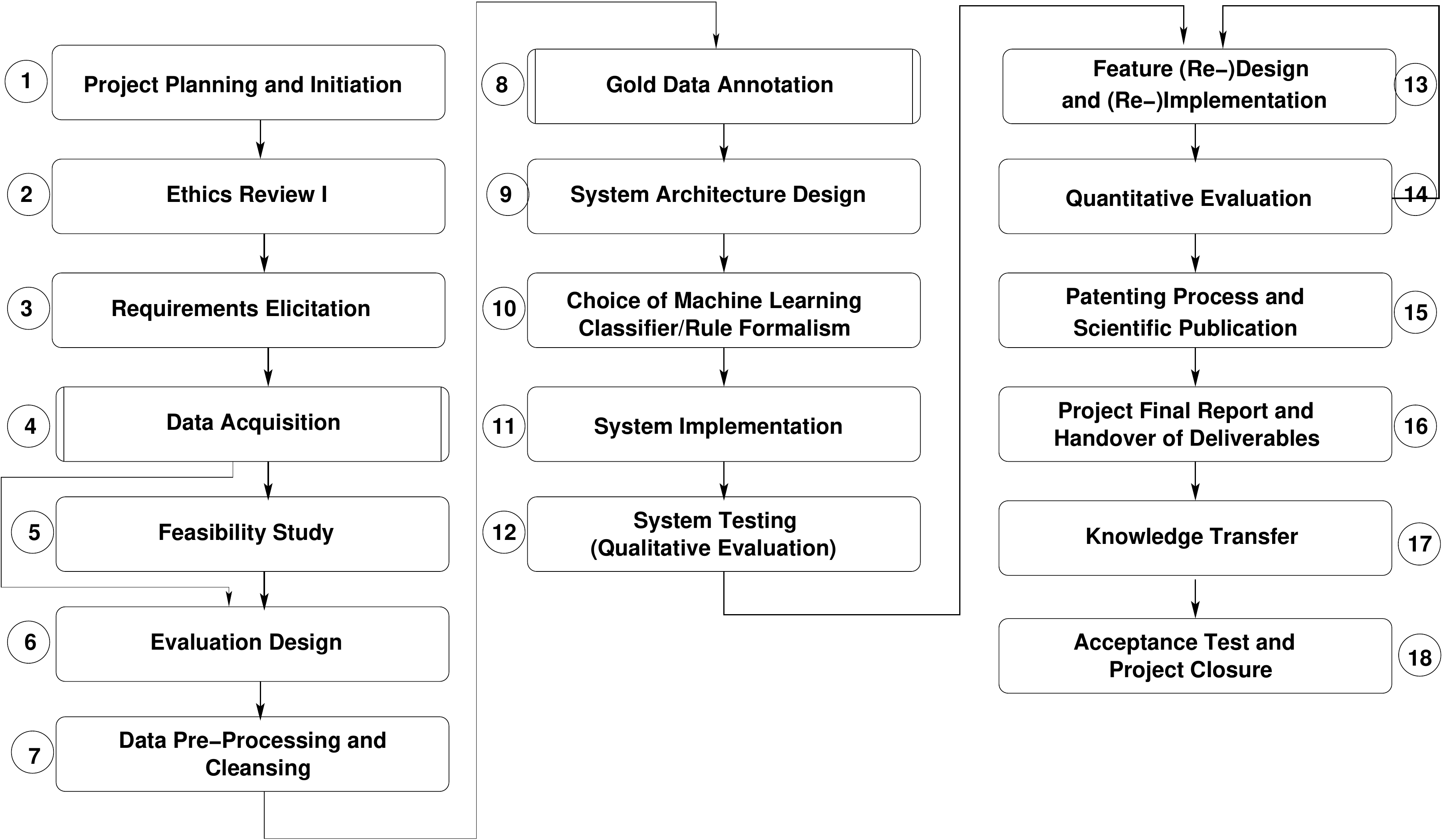}}
    \caption{Data-to-Value Process Model. Note two steps contain sub-processes.}
    \label{fig:d2v-process}
  \end{center}
\end{figure}

Figure~\ref{fig:d2v-process} shows the high-level process model behind
the Data-to-Value development methodology: in the beginning, the
Project Planning and Initiation stage is used to draft a project
charter, and to formally launch the project. A significant part of the
planning process is the specification of success, the planning of
automatic or manual evaluation procedures, and the budgeting of
evaluation resources.

\sloppypar{}
Once a business case is made (i.e., the value of a new product, service or
feature has been established, the \prjplan{} phase aims to author a project charter
and project plan, and initiate the project formally.
Before, after, or during this activity, an \ethicsreviewi{} is conducted to
answer the question whether the project and its output are morally objectionable.
Assuming no ethical obstacles, the \requirementselicitation{} phase seeks to obtain more detailed formal
requirements, both functional and non-functional.
In the \dataacquisition{} phase, autorization and access to any prerequisite
data-sets are obtained. 
The \feasibilitystudy{} phase, shown as a ``sub-process'' because it could be
seen as a light-weight version of the Figure~\ref{fig:d2v-process} as a whole,
is used to re-risk the project, by conducting some preliminary experiments on
a static data sample drawn from the full data-set(s) to be used by the
project. Although optional, it is highly recommended for all complex projects.
Very early in the process, in the \evaldesign{} phase, an experimental design
is worked out and an evaluation protocol is committed to. 
In the \preproc{} phase, all data sets are brought in
the right formats suitable for the project. In particular, eliminating noise
elements (data irrelevant to the project, or data that may be relevant that
violates formatting specifications), translating elements into a canonical
form, and linking data sets together where appropriate happens here. In some
projects, 80\% of the project time gets spent in this phase, and it is not
uncommon that this phase needs to be repeated due to the late recognition that
more work is required.

\begin{figure}
  \begin{center}
    {\includegraphics[width=.8\textwidth]{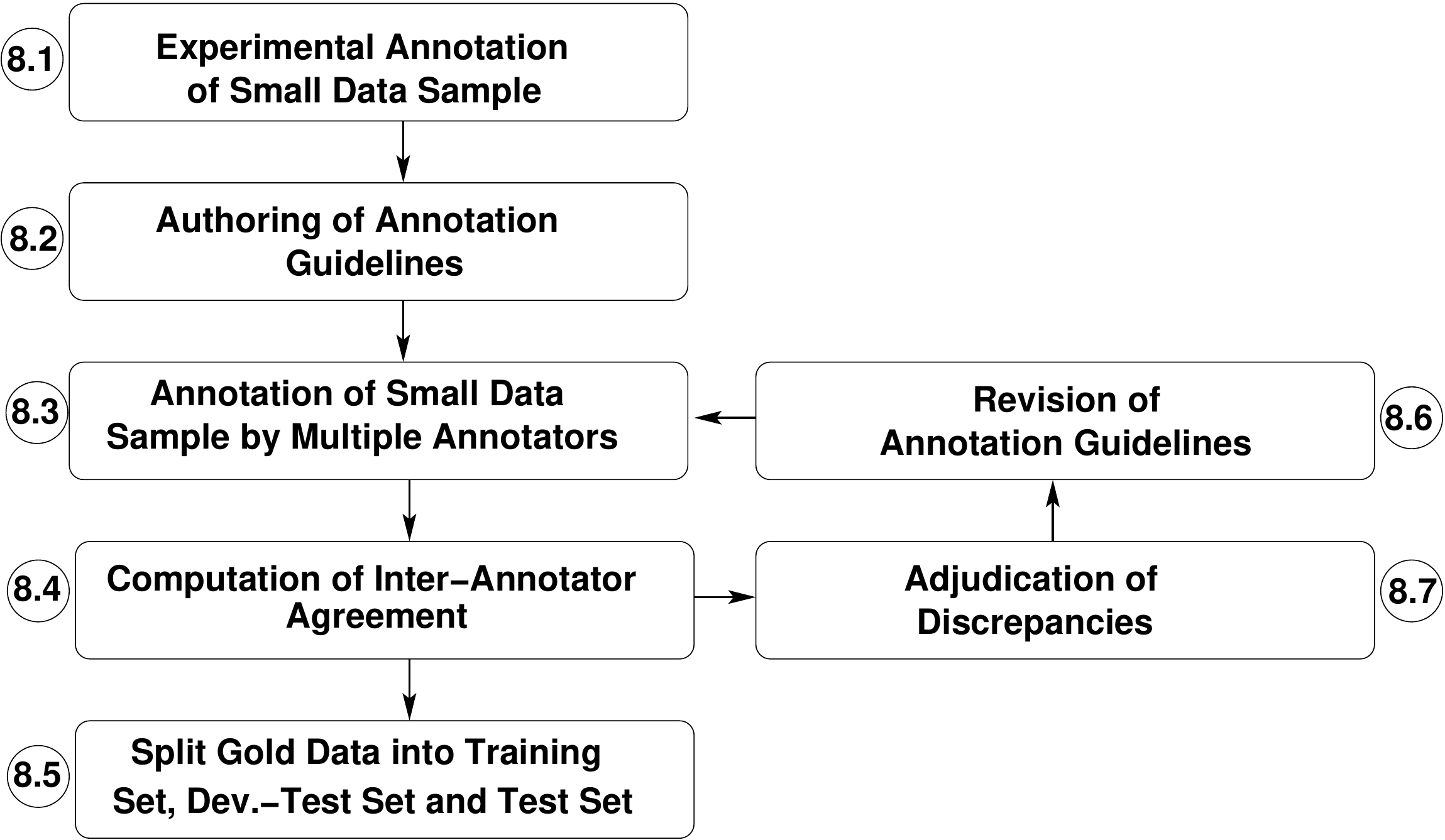}}
    \caption{Gold Data Annotation: A Sub-Process of D2V.}
    \label{fig:gold-standard-annotation}
  \end{center}
\end{figure}

\sloppypar{}
The \golddataannotation{} phase is a sub-process (shown in
Figure~\ref{fig:gold-standard-annotation} and explained below) dedicated to
creating data that gets used to train classifiers if a supervised training
regime is used; a second purpose is to create a ground truth for regular
automatic evaluation runs.


\sloppypar{}
The annotation of gold data requires care, patience, resources and foresight,
and if all of these are committed, the medium-term payback can be huge. In
contrast, if conducted sloppily or omitted, projects can fail entirely or at
least significant waste is likely to occur.
Figure~\ref{fig:gold-standard-annotation} shows the process, which typically
takes a few weeks or months, divided into multiple iterations. First, a
small, representative sample of the data to be annotated must be extracted from the
full data set. A few documents are inspected and informally annotated (\preannotate{}
phase). Once a particular style of annotation has been decided on, the \authorguidlines{}
phase can commence. The need for written guidelines is motivated by the
desire to install an objective process that relies exclusively on a document
artifact, and no longer on an individual's mental representation, which is
subjective. A larger data sample is then annotated. 5-10\% of it is processed by multiple 
annotators working on the same data points in an overlapping fashion
($k\ge{}3$) to compute Inter-Annotator Agreement (IAA) in the \agreement{}
phase) in order to measure the degree to which the task has been objectively specified. 
IAA is very important also because it defines the upper bound for the machine's
performance of the task (not 100\%!). The remaining 90-95\ can be processed by single annotators
unless the project is very well resourced and/or quality demands are highest. A gold 
data corpus can be created by manually adjudicating (\adjudicate{}) any discrepancies caused
by disagreeing annotators; alternatively;
if an odd value or $k$ was chosen, the ``truth'' can be selected from parallel 
annotator judgments using majority voting. Insufficent IAA leads to a
\reviseguidelines{} followed by re-annotation. Eventually, the gold data
corpus is split into three parts (\splitcorpus{} phase), a training set (``train'', often 60\% or
80\% of the annotated sample), a development test set (``dev-test'', often
10\%-20\%) and a test set (``test'', often 10\%-20\%).\footnote{
  The training set (optional, needed only
  when supervised machine learning is used) is used to induce (``learn'')
  models from. The dev-test partition is not looked at by humans, and only
  utilized for blind-run regular automatic evaluations (i.e., we look at the
  resulting evaluation scores, but we cannot investigate the data causing any
  errors). The test set is not used at all during development. Its use is
  confined to a single evaluation run at the end of the project, to define the
  definitive system quality.
}

\sloppypar{}
After that, the \sysdesign{} phase produces the architecture of the software
system. Then a particular paradigm is chosen in the \framing{} phase: if the system
is going to be a rule based system, the kind of rule processing technology or
framework, and if machine learning is to be used, which type of model. The framing
in terms of choice of classifier or rule formalism may impact the detailed architecture,
so in practice, the \sysdesign{} and \framing{} phases are intertwined more often than
not.
Then the \sysimplementation{} phase targets the actual software development
of the processing engine, which comprises most of the software but not any
rules or features, which are separately devised, and often by different team
members.
The \systesting{} phase, which includes both manual tests and automated
(unit) tests, is dedicated to the qualitative evaluation; in other words,
either the system is working to specification or there are still known bugs.

\sloppypar{}
With entering the \features{} phase, a series of
iterations begins that ultimately only terminates for one of two reasons:
either the projected time dedicated to this phase is used up, or the targeted
quality level has already been reached. Sometimes we may want to stop earlier
if no further progress seems possible, and additional efforts show diminishing
returns. Each iteration begins with feature brainstorming sessions and 
includes studying data from the training set, implementation sprints, and 
\quantevali{} steps to try out new features and their effectiveness.

\sloppypar{}
Ultimately, the loop exits, and four concluding phases wrap up the project
(\patentingpub{} to document the work and secure
the intellectual property rights to novel invented methods, authoring the
\finalreport{}, handing over the deliverables in the form of data,
software and documentation and the \knowledgetransfer{} phase, which verbally
communicates the findings to stakeholders and ascertains a full understanding
at the receiving end; finally, a successful \acceptclosure{} leads to the
formal project closure).

\sloppypar{}
While technically, our model for conducting \emph{research} projects could end
(\research{}),
industrial real-life systems get used in production, and modern systems require
recurring activities to update statistical model after deployment. Therefore, it
is advisable to have a look at what happens around the launch:
before we launch the system in the \deployment{} phase, we again conduct
an \ethicsreviewii{} to assess potential moral objections, this time
paying attention to morally questionable system functions and emergent
properties (e.g.~discrimination or unfairness).
Once the system is running, a \monitoring{} phase watches the system perform
its function whilst logging interesting events (system decisions, user activities).
At regular intervals, a \quantevalii{} phase is followed by \retraining{}, depending
on its findings, in order to keep statistical models ``fresh''.

\begin{figure}
  \begin{center}
    {\includegraphics[width=.8\textwidth]{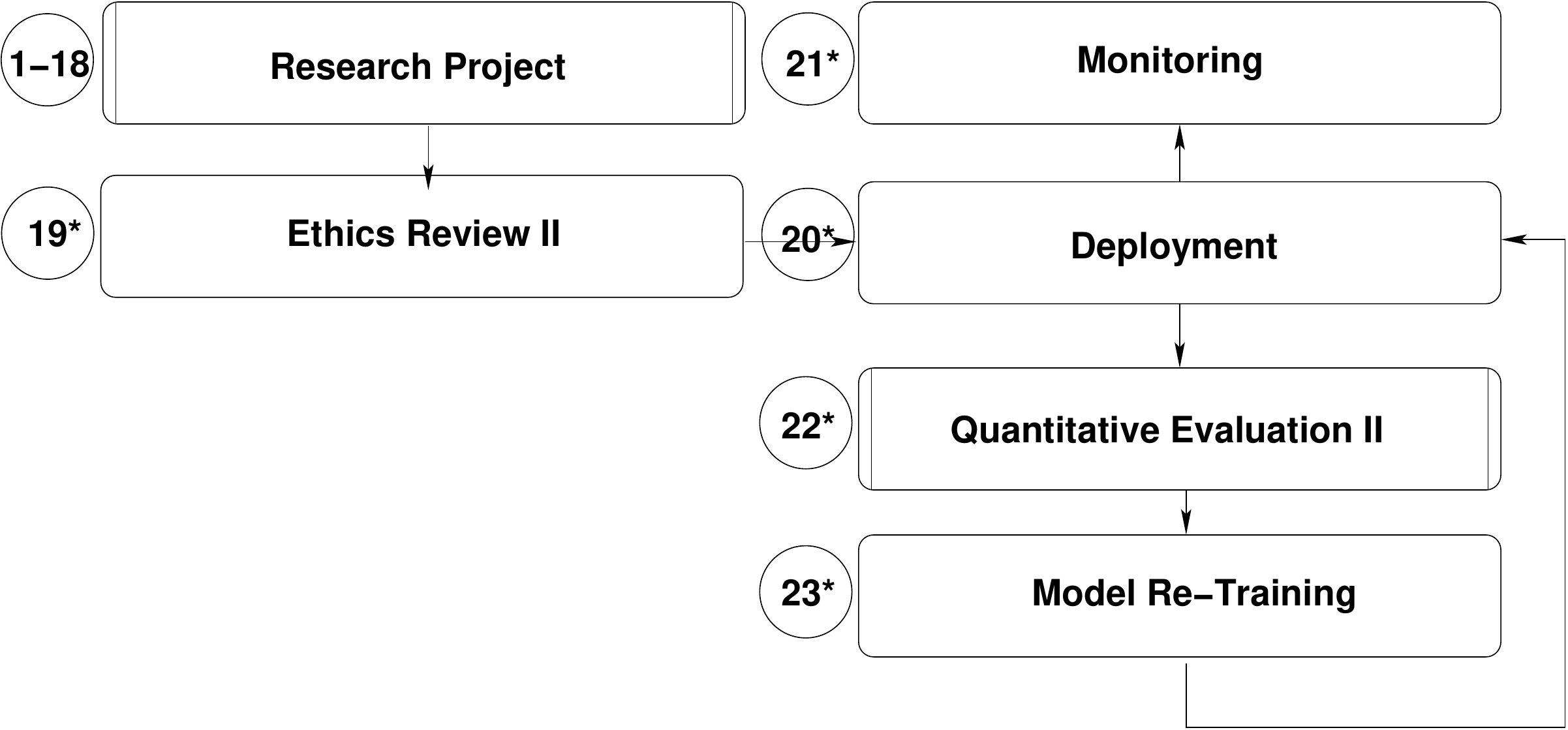}}
    \caption{From Development to Deployment and Operations. Note that if phases marked with an asterisk ($\ast$) are to be included, we are dealing with a research \& development project, not just a research project, and in that case you may need to move steps 16-18 after phase 23, or execute them twice.}
    \label{fig:deployment}
  \end{center}
\end{figure}

\sloppypar{}
Note that Figure~\ref{fig:d2v-process} shows only the process for
development of a system; in real-life settings, this is interleaved
with deploying and operating the result of the development process: a
piece of software gets sold as a product, uploaded to the Internet as
an open source offering, or deployed as a service (API) after
development (Figure~\ref{fig:deployment}). Upon launch, the longest
part of the software's life-cycle begins: it gets used by its target
audience, and it needs to be monitored (Monitoring phase) in order to
fix software defects. In the case of machine learning components,
ongoing quantitative evaluation is required, which either requires
grading system output post-hoc, or expanding the gold data with fresh
data samples to be annotated. If the Quantitative Evaluation suggest
the input data has deviated substantially from data the sytem was
trained on, Model Re-training is due. This cycle continues until the
product or service reaches its end of life.

\sloppypar{}
Each process phase has a set of guidance questions assigned to it
(cf.~Appendix~\ref{app:question-catalog} for some samples), which help
to increase project consistency. 

\section{Discussion} \label{sec:discussion}

The D2V methodology is quite a rich model, considering the number of
distinct phases. It is not aimed to be easily memorable, but was
designed to give the practitioner comprehensive guidance, which may
not be required for each project: a detailed question catalog
Experienced project managers will adjust the process to the complexity
and nature of project; for example, bigger or more complex projects
need more rigid processes and detailed formal documentation than small
studies conducted by teams of two.

A poll conducted twice within seven years in-between suggests that
CRISM-DM is consistently the most popular process by a large margin
(Table~\ref{tab:dm-methodology-stats}); that poll did not feature any
choices for unstructured or big-data specific processes at the time.
Only CRISP-DM has obtained broad adoption; however, neither of the
processes listed are actually potential substitutes for D2V, since
none of them have specific provisions for working with textual data.
Indeed, since most data falls into the unstructured category, it is
suprising that no process has previously been proposed.  One advantage
of adopting a new methodology is that it can be done immediately, and
any gains materialize immediately as well. This is in contrast to
technical choices (e.g.~programming languages) that often face a
chicken-and-egg problem in that there is a risk of adopting a
technology that does not already have a broad community.

\begin{table}
  \begin{center}
    \caption{An Internet poll conducted by the data mining portal \url{kdnuggets.com} in the years 2007 and 2014 (total $N=200$). Remarkably, relative popularity has not changed much in 7 years.} 
    \label{tab:dm-methodology-stats}
    \begin{tabular}{lrrl} \hline\hline
      Methodology & Used  & by & Respondents \\ \hline
      CRISP-DM    & 42\%  & -  & 43\% \\
      SEMMA       &  8\%  & -  & 13\% \\
      KDD         &  7\%  & -  & 8\% \\ \hline\hline
    \end{tabular}
  \end{center}
\end{table}

Table~\ref{tab:method-comparison} shows a summary comparison between
D2V and previous methodologies.  One limitation of D2V methodology is
that its dedication to unstructured projects makes it less suitable
for structured data mining projects, but that domain is sufficiently
addressed by the CRISP-DM model.  Another shortcoming of D2V, in the
version presented here, and also of all other models is that they do
not permit a prediction of the time spent in each phase, and of the
duration of the project overall; we will re-visit this in future work.

\begin{table}
  \begin{center}
    \caption{Evaluative Comparison between CRISP-DM, SEMMA, KDD and D2V: D2V provides more fine-grained phases and more substantial guidance with around one hundred process-supporting questions.}
    \label{tab:method-comparison}
    \begin{tabular}{lrllllll}\hline\hline
      Process       & Phases             & structured & unstructured & rule-based  & learning-based & guidance      & `evaluation- \\ 
      Model         &                    & data       & data         & approaches  & approaches     & questions     & first?'' \\ \hline
      CRISP-DM      & 6                  & yes        & no           & yes         & (yes)          & n/a           & no \\
      SEMMA         & 4-5                & yes        & no           & yes         & (yes)          & n/a           & no \\
      KDD           & 5-9                & yes        & no           & yes         & (yes)          & n/a           & no \\ 
      Marr Strat.~Board & 7              & yes        & no           & no          & no             & 6             & no \\ \hline
      \textbf{D2V}  & \textbf{30}        & \textbf{no}& \textbf{yes} & \textbf{yes}& $\;$\textbf{yes} & \textbf{96} & \textbf{yes} \\ \hline\hline
    \end{tabular}
  \end{center}
\end{table}

\section{Summary, Conclusion \& Future Work} \label{sec:conclusion}

In this paper, we have described a new process model for the
systematic pursuit of big data projects, in particular dedicated to
working with textual data. The proposed model, D2V, is different from
previous work in that is is not concerned with data mining (as
CRISP-DM, SEMMA and KDD are); instead, supervised learning of textual
structures are the main focus. It can be characterized as
``evaluation-first'', not just because it de-risks projects by
prioritizing the scrutinizing of success criteria, but also because it
includes provisions for gold standard annotation and an overall
iterative approach that terminates based on diminishing returns
informed by repeat evaluations. It is informed by a catalog of guiding
questions. We also discussed the advantages and shortcomings of the
D2V methodology. We believe this model has merit to improve awareness
of the best practices for professionals in projects using unstructured
data (natural language processing of text), in particular when supervised
machine learning is involved, which is increasingly common (information
extraction, sentiment analysis, document topic classification). The
D2V methodology can also aid the teaching of data science.

In future work, data collection exercises should be attempted to
measure typical absolute and relative resources spend in each phase,
in order to permit forecasting-oriented modeling towards cost and
quality estimation. Another avenue for future work is the predictive
modeling of quality as it relates to time and cost.

\begin{appendix}
\section{D2V Question Catalog (Excerpt)} \label{app:question-catalog}

This appendix lists a representative sample of questions from the D2V
catalog of questions (from a total of $N\approx{}100$ questions). 
The the full list of question will be included in a software tool
supporting the process introduced here, which is left for future work.

\parindent0em 
\begin{tabular}{lll}\hline\hline
  No.& Sample Question                                          & Area \\
  \hline
  Q9 & How correct, truthful, reliable and complete is the data & Veracity \\
     & in the data set?                                         & \\
  Q10& How quickly does the data grow (in byte/s)?              & Velocity \\
  Q37& How structured/formalized is the data?                   & Data Management \\
  Q46& What are the hypotheses that could be tested using this  & Value\\
     & data set?                                                & \\
  Q51& What workflow is this data part of (in my organization,  & Workflow\\
     & at my customers' sites)?                                 & \\
  Q65& Is it morally right to build the planned application?    & Ethics\\
  Q67& What licensing entitlements apply to the data set under  & Legal\\
     & consideration?                                           & \\
  Q72& Will the system to be built need to support multiple     & Linguistics\\
     & languages?                                               & \\
  Q74& Whose responsibility is the ongoing re-training of any   & Governance\\
     & machine learning models post-deployment?                 & \\
     \hline
\end{tabular}

\end{appendix}

\bibliography{d2v-methodology}
\bibliographystyle{splncs03}

\end{document}